\documentclass[runningheads]{llncs}
\usepackage{graphicx}

\usepackage{tikz}
\usepackage{comment}
\usepackage{amsmath,amssymb} 
\usepackage{color}

\usepackage[accsupp]{axessibility}  

\usepackage{epsfig}
\usepackage[normalem]{ulem}
\usepackage{algorithm}
\usepackage{algorithmic}
\usepackage{bm}
\usepackage{soul}
\usepackage{makecell}
\usepackage{dblfloatfix}    
\usepackage{xcolor} 
\usepackage[flushleft]{threeparttable} 
\usepackage{multirow} 
\usepackage{wrapfig,lipsum,booktabs} 

\usepackage{amsfonts}       
\usepackage{nicefrac}       
\usepackage{microtype}      
\usepackage{caption}
\usepackage{mathtools}
\usepackage{subcaption}
\usepackage[pagebackref=true,breaklinks=true,colorlinks,bookmarks=false]{hyperref}

\definecolor{Note_color}{rgb}{0.0, 0.0, 1.0}

\definecolor{cl114_color}{RGB}{0,150,150}

\definecolor{ForestGreen}{RGB}{34,139,34}
\definecolor{applegreen}{rgb}{0.55, 0.71, 0.0}

\renewcommand\footnotemark{}
\begin{document}
\pagestyle{headings}
\mainmatter
\def\ECCVSubNumber{320}  

\newcommand{\PaperTitle}{INGeo: Accelerating \ul{I}nstant \ul{N}eural Scene Reconstruction \\ with Noisy \ul{Geo}metry Priors}

\newcommand{\FrameworkName}{INGeo}
\title{\PaperTitle} 

\titlerunning{\FrameworkName}
%
\author{Chaojian Li\inst{1*\dagger} \thanks{* Equal contribution. $\dagger$ Work done while interning at Meta Reality Labs.} \and
Bichen Wu\inst{2*} \and
Albert Pumarola\inst{2}\ \and Peizhao Zhang\inst{2} \and \\ Yingyan Lin\inst{1} \and Peter Vajda\inst{2}}
\authorrunning{C. Li et al.}
%
\institute{Georgia Institute of Technology \and Meta Reality Labs
\\
\email{\{cli851, celine.lin\}@gatech.edu, \{wbc, apumarola, stzpz, vajdap\}@meta.com}}
\maketitle

\begin{abstract} 
We present a method that accelerates reconstruction of 3D scenes and objects, aiming to enable instant reconstruction on edge devices such as mobile phones and AR/VR headsets. While recent works have accelerated scene reconstruction training to minute/second-level on high-end GPUs, there is still a large gap to the goal of instant training on edge devices which is yet highly desired in many emerging applications such as immersive AR/VR. To this end, this work aims to further accelerate training by leveraging geometry priors of the target scene. Our method proposes strategies to alleviate the noise of the imperfect geometry priors to accelerate the training speed on top of the highly optimized Instant-NGP. On the NeRF Synthetic dataset, our work uses half of the training iterations to reach an average test PSNR of $>$30.

\end{abstract}

\begin{figure*}[!t]
  \centering
\includegraphics[width=1.0\linewidth]{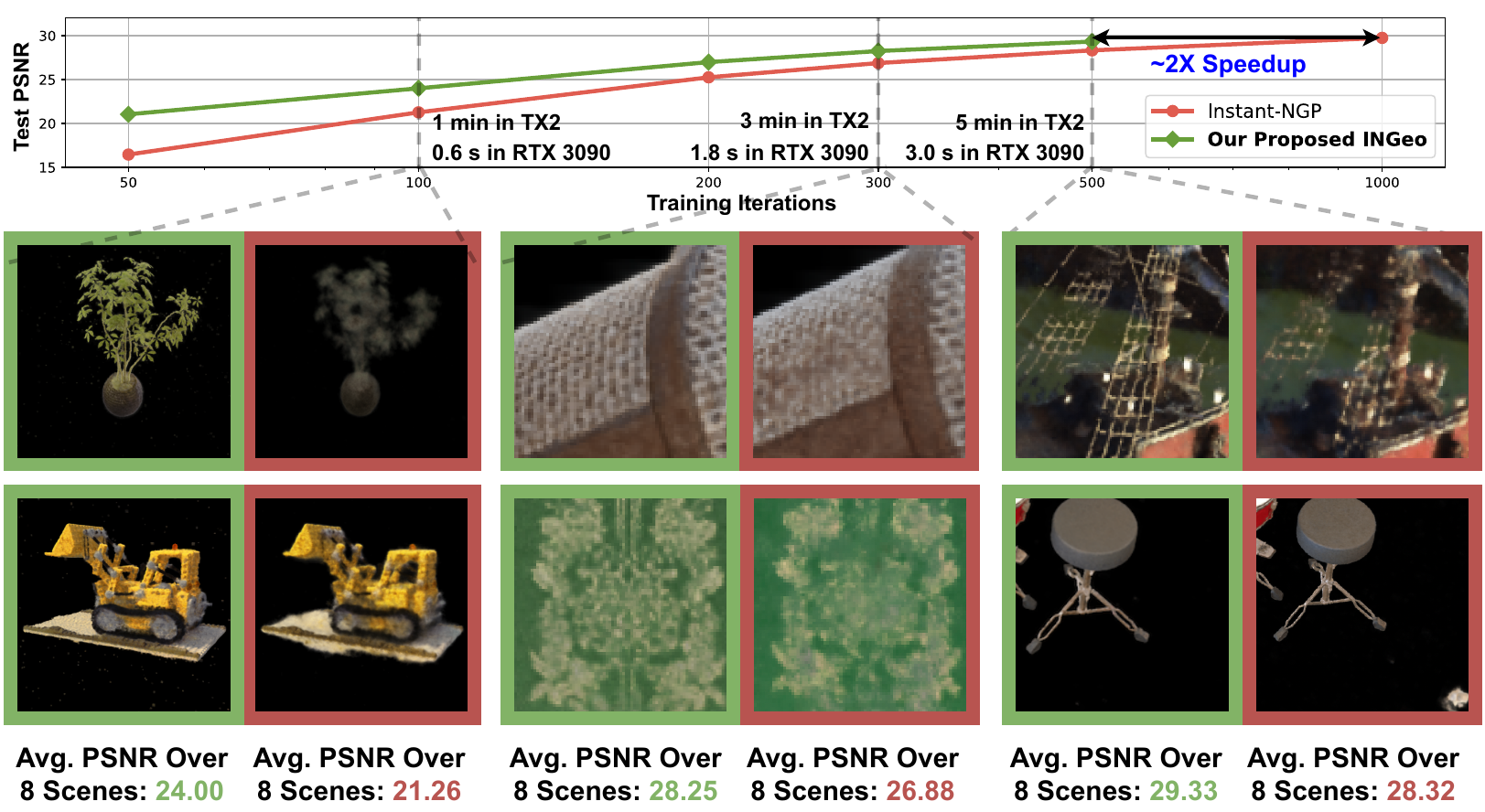}
\vspace{-1em}
\caption{{\FrameworkName} accelerates neural reconstruction training by $\sim$2$\times$ over the current SotA -- Instant-NGP \cite{muller2022instant}.
}
\label{fig:performace_overview}
\vspace{-1em}
\end{figure*}

\section{Introduction}
3D scene reconstruction has become a fundamental technology for many emerging applications such as Augmented Reality (AR), Virtual Reality (VR), and Mixed Reality (MR). In particular, many consumer-facing AR/VR/MR experiences (\textit{e.g.,} virtually teleporting 3D scenes, people, or objects to another user) require the capability to instantly reconstruct scenes on a local headset, which is particularly challenging as it requires both reconstruction quality and real-time speed.

Recently, neural rendering with implicit functions has recently become one of the most active research areas in 3D representation and reconstruction. Notably, Neural Radiance Field (NeRF) \cite{mildenhall2020nerf} and its following works demonstrate that a learnable function can be used to represent a 3D scene as a radiance field, which can then be combined with volume rendering \cite{max1995optical} to render novel view images, estimate scene geometry \cite{wei2021nerfingmvs,roessle2021dense,endres20133}, perform 3D-aware image editing such as re-lighting \cite{nerv2021}, and so on. Despite the  remarkable improvement in reconstruction quality, NeRF is extremely computationally expensive. For example, it requires $>$30 seconds to render an image and days of training time to reconstruct a scene on even high-end GPUs \cite{mildenhall2020nerf}. Recent works begins to focus on accelerating training of scene reconstruction. Specifically, one recent trend is to replace implicit function-based volume representations with explicit grid-based or point-based ones \cite{yu2021plenoxels,sun2021direct,chen2022tensorf,muller2022instant,xu2022point}. The advantage is that unlike implicit representation, where volume features are entangled, grid-based representations are spatially disentangled and thus the gradient updates at one location will not interfere with features that are far away \cite{muller2022instant}. Using grid-based representation, recent works have been able to accelerate the training of NeRF from days to minutes \cite{chen2022tensorf,yu2021plenoxels} to even seconds \cite{muller2022instant} on high-end GPUs. However, instant reconstruction on compute-constrained edge devices (mobile phones and AR/VR headsets) is still not possible.

This work sets out to \textit{reduce} the above gap, and boosts training efficiency by leveraging geometry priors to eliminate spatial redundancy during training. To do so, we leverage the intuition that most 3D scenes by nature are inherently sparse as noted by recent works \cite{muller2022instant,yu2021plenoxels,yu2021plenoctrees,chen2022tensorf,xu2022point,kondo2021vaxnerf}. Prior works~\cite{muller2022instant,chen2022tensorf,yu2021plenoxels,liu2020neural} have proposed to leverage a 3D occupancy grid to indicate whether a grid cell contains visible points or not. Such a occupancy grid is gradually updated based on the estimations of the volume density as training progresses.
Though this technique has been proven to be effective accelerating training in previous works \cite{muller2022instant,yu2021plenoxels,chen2022tensorf,liu2020neural}, obtaining a decent occupancy grid during training still takes a nontrivial amount of time, as discussed in Section~\ref{sec:method_pretrained_occupancy_grid}. Meanwhile, in many scenarios, as also noted by \cite{deng2021depth,xu2022point}, geometry of a scene, in the form of depth images, point-clouds, \textit{etc.}, can be obtained \textit{a priori} from many sources, such as RGB-D sensors and depth estimation algorithms (structure-from-motion, multi-view stereo, \textit{etc.}). These geometry priors can be converted to an occupancy grid that can be used in the reconstruction training. Despite that they are sparse and noisy, we hypothesize that with appropriate strategies to mitigate the impact of the noise, we can leverage such noisy geometry priors to further accelerate training.

\section{Related Works}
\textbf{Grid-based volume representation:} As discussed in \cite{muller2022instant}, one disadvantage of NeRF's implicit function based scene representation is that spatial features are entangled. In comparison, many recent works~\cite{sun2021direct,yu2021plenoxels,yu2021plenoctrees,chen2022tensorf,muller2022instant} that successfully accelerated training have switched to explicit grid-based representations, where volume features are spatially disentangled. To reduce the cubic complexity of the grid representation, these works adopted different compression techniques such as tensor factorization, hashing, multi-resolution, sparsifying voxels, and so on.

\begin{wrapfigure}{r}{0.55\linewidth}
\vspace{-2em}
\begin{subfigure}{\linewidth}
\includegraphics[width=1.0\linewidth]{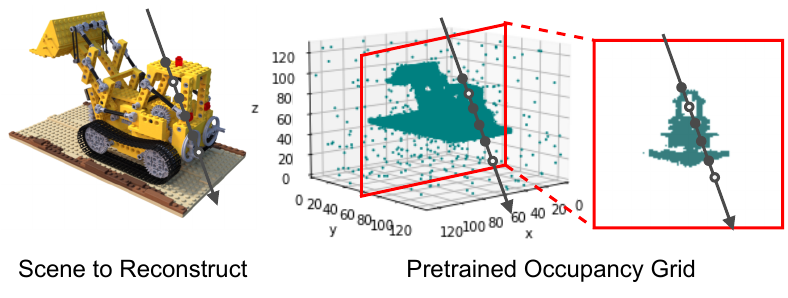}
\vspace{-2em}
\caption{}
\end{subfigure}


\begin{subfigure}{\linewidth}
\includegraphics[width=1.0\linewidth]{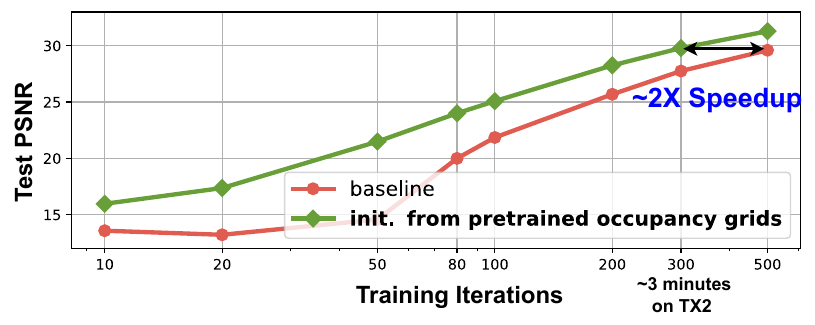}
\vspace{-2em}
\caption{}
\end{subfigure}
\vspace{-1em}
\caption{Visualization of the pretrained occupancy grid in (a) suggests the reason why it can accelerate the training process by $\sim$ 2$\times$ in (b) is that the occupancy grid can eliminate spatial redundancy in training.}
\label{fig:vis_well_trained_snapshot}
\vspace{-1.5em}
\end{wrapfigure}

\textbf{Using Geometry priors in NeRF:} Previous works have also tried to utilize geometry priors to boost the performance of NeRF in terms of training, rendering, and geometric estimation \cite{deng2021depth,neff2021donerf,xu2022point,roessle2021dense,roessle2021dense,wei2021nerfingmvs}. Depth-Supervised NeRF \cite{deng2021depth} utilized sparse depth estimation as an extra supervision in addition to RGB. It observed faster training and better few-shot performance. 
PointNeRF \cite{xu2022point} directly utilizes prior point-cloud to represent a radiance field. It enjoys the benefit of avoiding sampling in empty space and achieved good speedup over NeRF.

To \textbf{summarize} the relationship between our work and prior works: our work is built on top of the Instant-NGP \cite{muller2022instant}, the current record holder of the fast training, and we further accelerated it. We convert noisy geometry prior to an occupancy grid to reduce spatial redundancy during training and mitigate hash collision of Instant-NGP. This is more straightforward than Depth-Supervised NeRF \cite{deng2021depth}, which use prior depth as training signals. 

\section{Method}

\subsection{Can pretrained occupancy grids accelerate training?}
\label{sec:method_pretrained_occupancy_grid}

To obtain the occupancy grid, previous works~\cite{muller2022instant,chen2022tensorf} typically sample points in the 3D volume, compute densities, and update the occupancy grid. In practice, this boot-strapping strategy (iteratively update the occupancy grid and volume representation) usually converges well and can accelerate the training by concentrating training samples around critical regions of the scene. However, at the start of the training, we do not have a reliable occupancy grid, and we have to spend computational budgets to obtain one, which can take a non-trivial amount of time (e.g., it takes about 30 minutes on TX2 to obtain an occupancy grid with 88 \% IoU as compared to the final one on Lego scene~\cite{mildenhall2020nerf}). Thus, reducing the training time spent on learning the occupancy grid is critical for instant reconstruction of the scene.

We hypothesize that if an occupancy grid is provided \textit{a priori}, we can further accelerate training. To verify this, we conduct an experiment to re-train an Instant-NGP model but with its occupancy grid initialized from a pretrained model. The visualization of the pretrained occupancy grid can be found in Figure~\ref{fig:vis_well_trained_snapshot} (a).
As shown in Figure~\ref{fig:vis_well_trained_snapshot} (b), even though we simply inherited the occupancy grid from a pretrained model, while the majority of the model weights (i.e., 89\% even if we use floating point numbers to represent the dense occupancy grid) are randomly initialized, we still observed significant training speedup -- about 2x faster to reach an average of 30 PSNR on the NeRF synthetic dataset \cite{mildenhall2020nerf}. 

\subsection{Initialize occupancy grids with geometry priors}
\label{sec:method_geometry_prior}

\begin{wrapfigure}{r}{0.5\textwidth}
\vspace{-2.5em}
  \centering
  \includegraphics[width=1.0\linewidth]{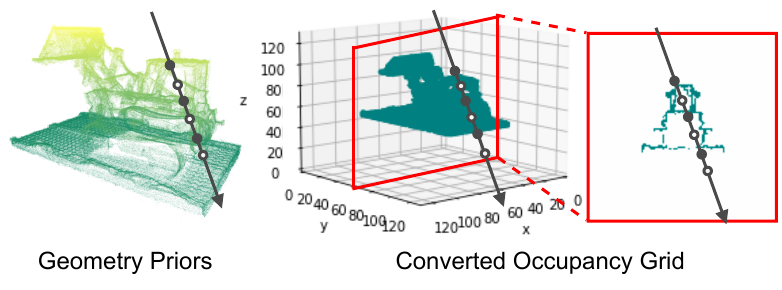}
\vspace{-2em}
\caption{Visualization of the point-cloud obtained by COLMAP~\cite{schoenberger2016sfm} and the converted occupancy grid.}
\label{fig:vis_colmap}
\vspace{-2em}
\end{wrapfigure}

As noted in \cite{deng2021depth,xu2022point}, in many application scenarios, such as on AR/VR headsets, geometry priors of a 3D scene can be obtained from many sources, such as RGB-D cameras and depth-estimation~\cite{schoenberger2016sfm,schoenberger2016mvs}. They can be converted to an occupancy grid of the scene. To illustrate this, we use the Lego scene from the NeRF-Synthetic dataset~\cite{mildenhall2020nerf} as an example to obtain a point-cloud using COLMAP \cite{schoenberger2016sfm,schoenberger2016mvs} and convert it to an occupancy grid, as shown in Figure~\ref{fig:vis_colmap}.

We notice that the occupancy grid obtained from point-cloud is similar to the one obtained from the pretrained model. We also notice missing points in the point-cloud and missing regions in the occupancy grid (e.g., bottom board below the bulldozer in Figure~\ref{fig:vis_colmap}), which could have an negative impact on the scene reconstruction.

So the question is how to design strategies to mitigate the impact of the noise and leverage the noisy geometry priors to accelerate training. Our strategy can be described as the following three aspects: density scaling, point-cloud splatting, and updating occupancy grid.

\begin{wrapfigure}{r}{0.6\textwidth}
\vspace{-2.5em}
\begin{subfigure}{\linewidth}
\includegraphics[width=1.0\linewidth]{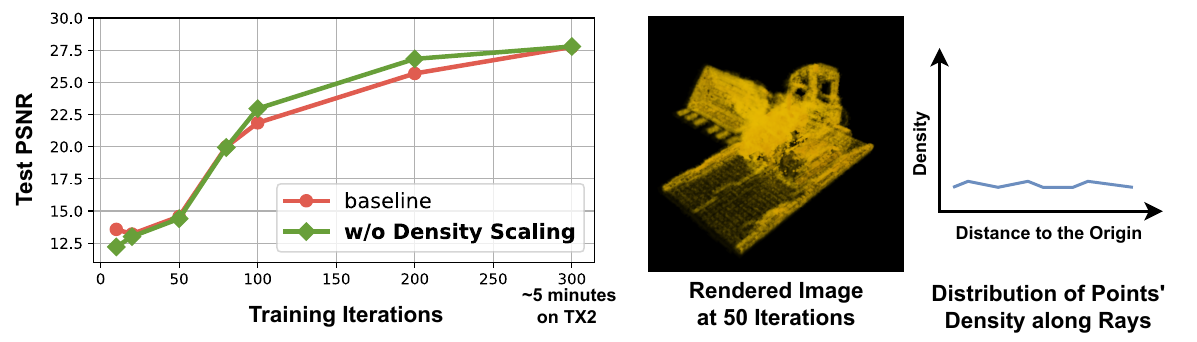}
\vspace{-2em}
\caption{}
\end{subfigure}


\begin{subfigure}{\linewidth}
\includegraphics[width=1.0\linewidth]{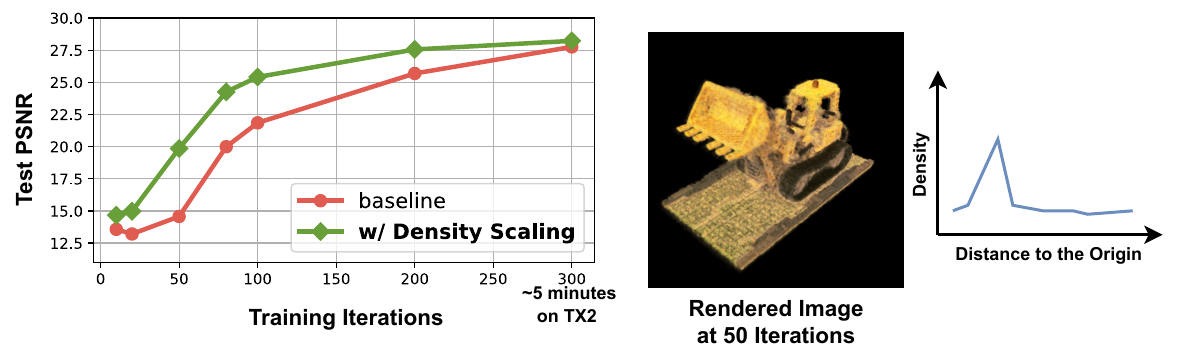}
\vspace{-2em}
\caption{}
\vspace{-1em}
\end{subfigure}
\caption{Comparing the effectiveness of geometry priors (a) w/o and (b) w/ the proposed \textbf{density scaling} on the Lego scene~\cite{mildenhall2020nerf}. Compare their training speed curves, rendered images at 50 iterations, and density distribution along a ray.}
\label{fig:density_scaling}
\vspace{-1em}
\end{wrapfigure}

\textbf{Density scaling}: When we first load an initialized occupancy grid to the model, to our surprise, we did not observe obvious speedup over the baseline using a random occupancy grid, as shown in the trainign speed curve in Figure~\ref{fig:density_scaling} (a). After investigation, we found that the reason can be explained as the following: an initialized occupancy grid can ensure we sample and train points with the positive grid cells. However, the initial density values inside the cells are computed by the model, whose weights are randomly initialized. At the beginning of the training, the density prediction of the model is generally low. As an evidence, see the rendered image in Figure~\ref{fig:density_scaling} (a), note that the image looks quite transparent. Also note the density along the ray is generally low. 

This is problematic, since in a well-trained model, densities around the object surface are generally much higher. This means that the accumulated transmittance will quickly drop to 0 after the ray encounters the surface. As a result, during back-propagation, the gradient of the color prediction will mainly impact the surface points. However, if the density along the entire ray is low, the color gradient will broadcast to all the sampled points along the ray, even for those that should be occluded by the surface. This slows down the training.

\begin{wrapfigure}{r}{0.6\textwidth}
\vspace{-2.5em}
\begin{subfigure}{\linewidth}
\includegraphics[width=1.0\linewidth]{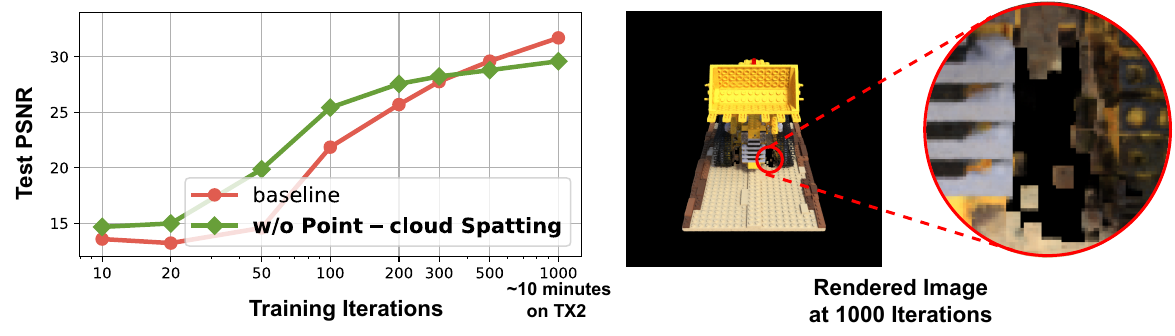}
\vspace{-2em}
\caption{}
\end{subfigure}


\begin{subfigure}{\linewidth}
\includegraphics[width=1.0\linewidth]{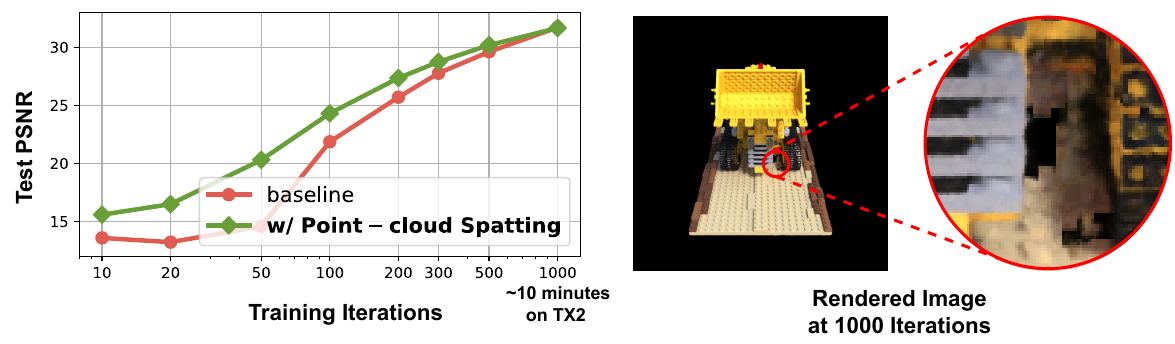}
\vspace{-2em}
\caption{}
\end{subfigure}
\vspace{-1em}
\caption{Compare (a) w/o (i.e., only w/ the proposed density scaling) and (b) w/ the proposed \textbf{point-cloud splatting} on Lego dataset~\cite{mildenhall2020nerf} in terms of the training efficiency over baselines (e.g., training from random initialization) and rendered images.}
\label{fig:point_cloud_spatting}
\vspace{-2em}
\end{wrapfigure}

To fix this, we decide to include an inductive bias to make sure that rays terminate soon after it encounters the object surface that is identified by the occupancy grid. This can be implemented by simply scale up the initial density prediction by a factor. Our experiments show that this made the densities at surface points much larger, as shown in  Figure~\ref{fig:density_scaling} (b), the rendered images becomes more ``solid'', and we also observed more significant speedup over the baseline.

\textbf{Point-cloud splatting}: When converting a point-cloud to an occupancy grid, the default option is to set the element of the occupancy tensor to positive if a point is within its corresponding cell. However, point-clouds are typically sparse and incomplete, and the corresponding occupancy grid also contains a lot of missing regions. After adopting the density scaling, despite the significant performance gain at the initial iterations of the training, we notice that the PSNR plateaus very soon. Visualization of the rendered images show missing regions, as a result of incomplete occupancy grid, see the zoom-in in Figure~\ref{fig:point_cloud_spatting} (a). To address this, one simple strategy is to not to treat each point as infinitesimal, but let it splat to nearby grid cells within a radius.  This can help complete small missing regions, at the cost of adding more empty grid cells. We used this strategy to initialize the occupancy grid, and notice significant performance improvement both quantitatively and qualitatively, as shown in Figure~\ref{fig:point_cloud_spatting} (b), although it still cannot completely remove the mission region.

\begin{wrapfigure}{r}{0.6\textwidth}
\vspace{-2em}
\begin{subfigure}{\linewidth}
\includegraphics[width=1.0\linewidth]{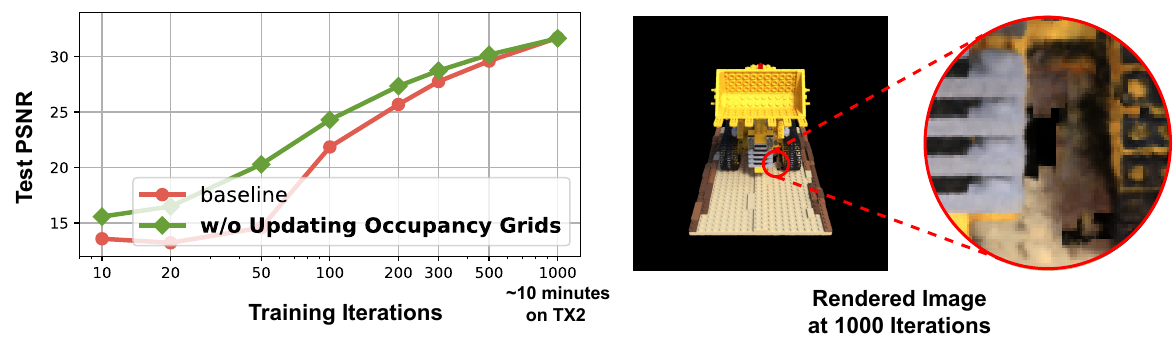}
\vspace{-2em}
\caption{}
\end{subfigure}


\begin{subfigure}{\linewidth}
\includegraphics[width=1.0\linewidth]{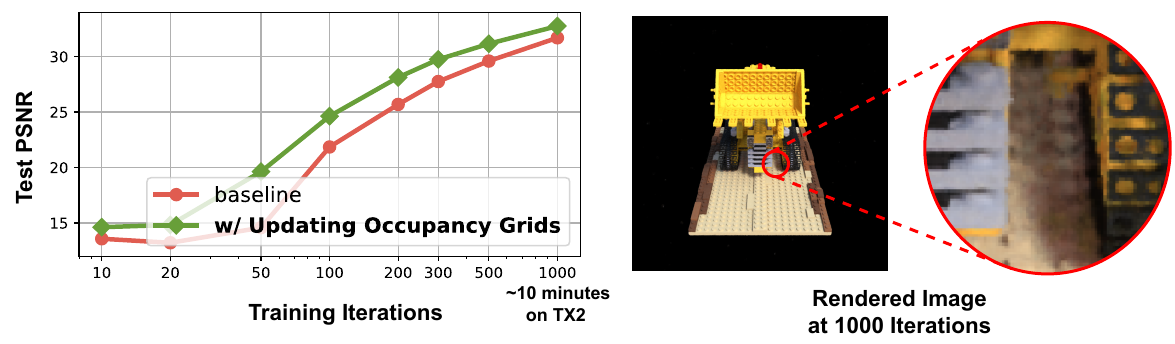}
\vspace{-2em}
\caption{}
\end{subfigure}
\vspace{-1em}
\caption{Compare (a) w/o (i.e., only w/ the proposed density scaling and point-cloud splatting) and (b) w/o the proposed \textbf{updating occupancy grids} on Lego dataset~\cite{mildenhall2020nerf} in terms of the training efficiency over baselines (e.g., training from random initialization) and rendered images.}
\label{fig:update_occupancy_grids}
\vspace{-2em}
\end{wrapfigure}

\textbf{Updating occupancy grids}: 
Sparsity and errors in the geometry prior are inevitable and we cannot completely rely on the initial occupancy grid throughout the training. To address this, we adopt a new strategy to combine initialized occupancy grid with continuously updated ones. More specifically, we start with an initialized occupancy grid, but also follow previous works~\cite{muller2022instant,chen2022tensorf}
to re-evaluate volume density at a certain frequency, and update the occupancy grid.  We assume that the initial occupancy grid obtained from geomery priors (depth sensor, depth-estimation) are sparse but accurate (low-recall but high precision). Therefore, we always mark the initially occupied cells as positive, and add new cells whose estimated density are larger than a threshold. This effectively addressed the missing point issue, as shown in Figure~\ref{fig:update_occupancy_grids}.

\section{Experiments}
We run experiments to verify whether noisy geometry priors can further accelerate the training of scene reconstruction on top of the highly optimized Instant-NGP. We first summarize our experimental settings in Section \ref{sec:exp_setting} and report the training speed comparison in Section~\ref{sec:compare_with_sota}. 

\subsection{Experiments settings}
\label{sec:exp_setting}

\textbf{Models and datasets.} Our implementation is based on the open-sourced Instant-NGP \cite{muller2022instant}. Following the settings in~\cite{muller2022instant}, our model uses 16 hash tables, each has $2^{19}$ entries and 2 features per entry. Besides, a 1-layer MLP and a 2-layer MLP are used to predict the density and color, respectively. The resolution of the occupancy grid is $128^3$ for all scenes.  Same as~\cite{muller2022instant}, we benchmark ours proposed {\FrameworkName} on the commonly used NeRF-Synthetic dataset~\cite{mildenhall2020nerf}  with 8 scenes.

\textbf{Baselines and evaluation metrics.} We select the highly optimized Instant-NGP~\cite{muller2022instant} as the baseline. To qualify the training efficiency, we use the number of \ul{training iterations} and \ul{training time} as the metrics for training cost, and PSNR on the test set as the metrics for reconstruction quality. The training time is measured on an embedded GPU, NVIDIA Jetson TX2~\cite{tx2}. We did not include other works as baseline, since Instant-NGP is significantly faster than all the existing methods that we are aware of.  

\begin{table*}[!t]
\caption{Comparing our propose {\FrameworkName} with the SotA efficient training solution, Instant-NGP~\cite{muller2022instant}, in terms of the achieved PSNR under given training cost (i.e., training iterations or training time).}
\centering
  \resizebox{1.0\linewidth}{!}
  {
    \begin{tabular}{c|c|c||cccccccc|c}
    \toprule
    \multirow{2}{*}{Method} & \multirow{2}{*}{\# Iter.} & \multirow{2}{*}{Time on TX2} & \multicolumn{9}{c}{Test PSNR} \\
     & & & Mic & Ficus & Chair & Hotdog & Materials & Drums & Ship & Lego & Avg. \\
     \midrule
     Instant-NGP~\cite{muller2022instant} & \multirow{2}{*}{10} & \multirow{2}{*}{6 seconds} & 15.16 &14.85 &12.94 &14.41 &15.36 &13.04 &17.95 &13.57 &14.66 \\
\textbf{{\FrameworkName}} &  &  & 22.21 &20.54 &15.41 &15.97 &18.11 &15.77 &18.81 &14.61 & \textbf{17.68 ($\uparrow$ 3.02)} \\
     \midrule
     Instant-NGP~\cite{muller2022instant} & \multirow{2}{*}{100} & \multirow{2}{*}{1 minute} & 25.85 &20.26 &22.78 &24.01 &22.15 &18.12 &15.07 &21.85 &21.26\\
\textbf{{\FrameworkName}} &  &  & 27.94 &25.92 &24.00 &25.17 &22.02 &20.99 &21.33 &24.63 & \textbf{24.00 ($\uparrow$ 2.74)} \\
     \midrule
     Instant-NGP~\cite{muller2022instant} & \multirow{2}{*}{500} & \multirow{2}{*}{5 minutes} & 31.72 &29.17 &29.84 &33.01 &25.92 &23.37 &23.97 &29.60 &28.32 \\
\textbf{{\FrameworkName}} &  &  & 32.47 &29.55 &31.18 &33.24 &26.04 &23.71 &27.27 &31.16 & \textbf{29.33 ($\uparrow$ 1.01)} \\
     \midrule
     Instant-NGP~\cite{muller2022instant} & \multirow{2}{*}{1K} & \multirow{2}{*}{10 minutes} & 33.03 &30.27 &31.60 &34.28 &26.77 &24.27 &26.06 &31.70 &29.75 \\
\textbf{{\FrameworkName}} &  &  & 33.66 &30.41 &32.72 &34.41 &27.02 &24.49 &28.67 &32.76 &  \textbf{30.52 ($\uparrow$ 0.77)}\\
     \midrule
     Instant-NGP~\cite{muller2022instant} & \multirow{2}{*}{30K} & \multirow{2}{*}{5 hours} & 35.68 &32.34 &34.83 &36.99 &29.46 &25.85 &30.25 &35.60 &32.63 \\
\textbf{{\FrameworkName}} &  &  & 36.17 &31.54 &35.04 &36.98 &29.15 &25.82 &30.95 &35.47 & \textbf{32.64 ($\uparrow$ 0.01)} \\
    \bottomrule
    \end{tabular}
    }
  \label{tab:compare_with_sota}
  \vspace{-1em} 
\end{table*}

\subsection{Training speed comparison with SotA}
\label{sec:compare_with_sota}

As demonstrated in Figure~\ref{fig:performace_overview} and Table~\ref{tab:compare_with_sota}, we observe that (1) to reach an average test PSNR of 30 on the 8 scenes on the NeRF synthetic dataset, our method uses half of the training iterations than Instant-NGP. (2) Under different training budgets (from 6 sceonds to 10 minutes), our proposed {\FrameworkName} consistently achieves  $\uparrow$ 0.77 \% $\sim$ $\uparrow$ 3.02 \% higher average test PSNR than Instant-NGP. (3) With a much longer training budget (30K iterations, 5 hours on a TX2 GPU), the advantage of our work diminishes, but {\FrameworkName} does not hurt the performance, and is still slightly better ($\uparrow$ 0.01 PSNR) than Instant-NGP. Based on the observations above, we conclude that the proposed {\FrameworkName} makes it possible to achieve instant neural scene reconstruction on edge devices, achieving $>$ 30 test PSNR within 10 minutes of training on the embedded GPU TX2, and $>$ 24 PSNR within 1 minute of training. To check the individual effectiveness of our proposed noise mitigation techniques -- density scaling, point-cloud splatting, and updating occupancy grid, we refer readers to Section~\ref{sec:method_geometry_prior} and Figure~\ref{fig:density_scaling} $\sim$~\ref{fig:update_occupancy_grids}.

\section{Conclusion}
In this paper, we present {\FrameworkName}, a method to accelerate reconstruction of 3D scenes. Our method is built on top of the SotA fast reconstruction method, Instant-NGP, and we utilized geometry priors to further accelerate training. We proposed three strategies to mitigate the negative impact of noise in the geometry prior, and our method is able to accelerate Instant-NGP training by 2 $\times$ to reach an average test PSNR of 30 on the NeRF synthetic dataset. We believe this work bring us closer to the target of instant scene reconstruction on edge devices. 

\clearpage
%
%
\bibliographystyle{splncs04}
\bibliography{egbib}

\end{document}